\pdfoutput=1 

\documentclass[11pt]{article}

\usepackage[]{acl}

\usepackage{times}
\usepackage{latexsym}
\usepackage{bbm}
\usepackage[T1]{fontenc}

\usepackage[utf8]{inputenc}

\usepackage{microtype}
\usepackage{booktabs}
\usepackage{graphicx}
\usepackage{multirow}

\usepackage{graphicx}
\usepackage{caption}
\usepackage{subcaption}

\definecolor{mygreen}{HTML}{00786C}
\definecolor{lightgrey}{HTML}{447777}
\definecolor{mypurple}{HTML}{D64C1D}
\newcommand{\dataset}{\textsc{TyDiP}}
\newcommand{\data}{\textsc{TyDiP} }

\title{\textsc{TyDiP}: A Dataset for Politeness Classification in\\ Nine Typologically Diverse Languages }

\author{Anirudh Srinivasan ~~~ Eunsol Choi \\
        Department of Computer Science \\
        The University of Texas at Austin \\
        \texttt{\{anirudhs, eunsol\}@utexas.edu}}

\begin{document}
\maketitle

\begin{abstract}
We study politeness phenomena in nine typologically diverse languages. Politeness is an important facet of communication and is sometimes argued to be cultural-specific, yet existing computational linguistic study is limited to English. We create \dataset, a dataset containing three-way politeness annotations for 500 examples in each language, totaling 4.5K examples. We evaluate how well multilingual models can identify politeness levels -- they show a fairly robust zero-shot transfer ability, yet fall short of estimated human accuracy significantly. We further study mapping the English politeness strategy lexicon into nine languages via automatic translation and lexicon induction, analyzing whether each strategy's impact stays consistent across languages. Lastly, we empirically study the complicated relationship between formality and politeness through transfer experiments. We hope our dataset will support various research questions and applications, from evaluating multilingual models to constructing polite multilingual agents.\footnote{The data and code is publicly available at \url{https://github.com/Genius1237/TyDiP}.}

\end{abstract}

\section{Introduction}
Whether politeness phenomena and strategies are universal across languages or not have been controversial among sociologists and linguists. While ~\citet{brown1978universals} claimed their universality, other followup work~\cite{korac2001low} claimed how communication patterns can differ based on cultures and other social constructs such as gender~\cite{Mills2003GenderAP} and domains. 

To contribute to the linguistic study of cross-cultural politeness, we collect politeness labels on nine typologically and culturally diverse languages, Hindi, Korean, Spanish, Tamil, French, Vietnamese, Russian, Afrikaans, and Hungarian. This language set covers five scripts and eight language families. We follow the seminal work~\cite{danescu-niculescu-mizil-etal-2013-computational} closely, focusing on politeness exhibited in requests as they involve the speaker imposing on the listener, requiring them to employ various politeness techniques. To capture rich linguistic strategies that can be lost in translation~\cite{Lembersky2011LanguageMF}, we collect sentences written in each target language. To minimize the domain shift among languages, we collect examples in each language from their respective Wikipedia User talk pages, where editors make requests about administrative and editorial decisions. 


Crowdsourcing labels in low-resource languages is challenging. Thus, we carefully design an annotation process that includes a translation task to evaluate annotator's language proficiency and a model-in-the-loop qualification task which filters workers whose labels diverges from highly confident predictions from multilingual models. After this process, we observe high agreements among the annotators in our dataset despite the subjectivity of the task. Interestingly, the annotators agree with each other more when assigning politeness score on requests in their native languages compared to assigning politeness score on requests in English, which is their second language. 

Equipped with our new multilingual politeness dataset, we evaluate zero-shot transfer ability of existing multilingual models in predicting politeness -- subjective and pragmatic language interpretation task. Pretrained language models~\cite{conneau-etal-2020-unsupervised} fine-tuned on annotated English politeness data~\cite{danescu-niculescu-mizil-etal-2013-computational} show competitive performances on all languages, weighing in the universality of politeness phenomena across languages. We also witness impressive zero-shot performance of a high-capacity pretrained language model~\cite{NEURIPS2020_1457c0d6}. We observe a degradation in classification performances when we translate the target language (via Google Translate API) to English, suggesting politeness might not be preserved in the current machine translation model. Despite the simplicity of classification task, we report a substantial difference between the estimated human accuracy and the best model accuracy (over 10\% difference in accuracy in six out of nine languages). 

Lastly, we provide two studies delving into politeness phenomena. We map English politeness strategy lexicon to create politeness strategy lexicon in nine languages by using tools like automatic translation, lexicon alignment~\cite{dou-neubig-2021-word} and large-scale corpora in the same domain. Despite the limitations of automatic lexicon mapping, we largely observe consistent correlation with politeness score for each politeness strategy in nine languages we study, with some interesting exceptions. We then compare the notion of politeness and formality which has been studied in multilingual setting~\cite{briakou-etal-2021-ola}. Our empirical results supports that notions of politeness and formality cannot be used interchangeably. However, when we control for semantics, politeness classifier judges the formal version of the same sentence as more polite than its informal variant.



We release our annotated data and aligned politeness lexicon to support future work. Our dataset can support various end applications, such as building multilingual agents optimized for politeness~\cite{Silva2022PoliteTD}, developing a translation model that preserves politeness level~\cite{fu-etal-2020-facilitating}, evaluating the impact of different pretraining corpus and modeling architecture for modeling subjective tasks in a wide range of languages~\cite{Hu2020XTREMEAM}, understanding cultural-specific politeness strategies, and many more.
\section{\dataset: Multilingual Politeness Dataset}
\paragraph{Motivation}
Our goal is to construct high-quality multilingual evaluation data with native content, covering a wide range of languages on the task of politeness prediction. Following prior work~\cite{danescu-niculescu-mizil-etal-2013-computational}, we focus on identifying politeness in requests, where requests involve speaker imposing on the listener. This scenario elicit speakers to employ diverse strategies to minimize the imposition of requests, or apologizing for the imposition~\cite{lakoff1977you}. For each request text, we aim to collect a graded politeness score (between -3 and 3, with 0.5 increment).

\paragraph{Language Selection}
We chose Hindi, Korean, Spanish, Tamil, French, Vietnamese, Russian, Afrikaans, and Hungarian. Our criteria for selecting languages were (1) covering low resource language when possible, (2) languages with rich discussion on Wikipedia editor forum and (3) languages where we can recruit native speaker annotators on a crowdsourcing platform, Prolific.\footnote{\url{https://prolific.co/}}

\paragraph{Source Sentence Collection}
We source requests from Wikipedia user talk pages from target language Wikipedia dumps.\footnote{\url{https://dumps.wikimedia.org/backup-index.html}} Each request is a part of a conversation between editors on Wikipedia. We follow the pre-processing step from prior work~\cite{danescu-niculescu-mizil-etal-2013-computational}, extracting each request as a sequence of two successive sentences where the second sentence ends with a question mark ({?}). We present one example here: {"I'm somewhat puzzled by your recent edits on the Harper page, which have left two different sets of footnotes. Could you please explain your rationale for the change?"}

\begin{table*}[t]
    \small
    \centering
    \begin{tabular}{lllrrr} 
        \toprule
        Language        & Family        & Script     & Total \# Requests & \% Target / English / Other & Avg length (in bytes)  \\ 
        \midrule
        Hindi      (hi) & Indo-Aryan    & Devanagari & 4,412             & 71 / 26 / 3                 & 351                    \\
        Korean     (ko) & Korean        & Hangul     & 43,219            & 96 / 3 / 1                  & 183                    \\
        Spanish    (es) & Romance       & Latin      & 180,832           & 97 / 2 / 1                  & 181                    \\
        Tamil      (ta) & Dravidian     & Tamil      & 5,590             & 92 / 8 / 0                  & 325                    \\
        French     (fr) & Romance       & Latin      & 354,544           & 98 / 1 / 1                  & 179                    \\
        Vietnamese (vi) & Austroasiatic & Latin      & 22,070            & 96 / 4 / 0                  & 210                    \\
        Russian    (ru) & Slavic        & Cyrillic   & 291,220           & 98 / 1 / 1                  & 254                    \\
        Afrikaans  (af) & Germanic      & Latin      & 3,399             & 85 / 11/ 4                  & 134                    \\
        Hungarian  (hu) & Uralic        & Latin      & 80,825            & 98 / 1 / 1                  & 132                    \\
        \bottomrule
    \end{tabular}\vspace{-0.4em}
    \caption{Languages chosen for our study and their data statistics. We report the number of available requests in Wikipedia User talk pages after pre-processing step, the distribution of languages after language identification, and the average length in bytes for each request. } \vspace{-1em}
    \label{tab:langs}
\end{table*}

\subsection{Annotation Process}
Collecting annotations for non-English data for a wide range of languages is non-trivial in all aspects, from source text collection, annotator recruiting to annotation validation. We describe our annotation process here and hope that our collection strategy can provide insights for future multilingual data collection efforts for other tasks and domains. 

\paragraph{Pre-processing}
We observe that a sizable portion of the requests is written in language other than its own. Thus, we filter sentences not belonging to the target language with a language identification with \texttt{langdetect} \cite{nakatani2010langdetect}. 

Table \ref{tab:langs} shows data statistics, including the language distribution among these requests. We use the \texttt{Polyglot} tokenizer\footnote{\url{https://github.com/aboSamoor/polyglot}} for preprocessing.

\paragraph{Annotator Recruiting}

We collect our annotation on a crowdsourcing platform, Prolific, which allows us to find workers based on their first language. Instead of developing separate guidelines for each language, we recruit bilingual annotators. We also filter by their task approval rate ($> 98\%$). 

To annotators who meet these criteria, we perform qualification process which involves translation task and the target task, which we describe below.

\paragraph{Target Task Qualification}
Inspired by strong zero-shot transfer performances of multilingual models on a variety of tasks \cite{conneau-etal-2018-xnli, Wu2019BetoBB}, we use a multilingual classifier trained on existing English politeness dataset~\cite{danescu-niculescu-mizil-etal-2013-computational} to select sentences for the qualification task.\footnote{We describe this model (XLMR-target) in detail in Section 4.} We sample examples where the classifier assigned very high or very low politeness score for each language. Language-proficient researchers verified the correctness of model predictions on a subset (four) of languages. While the model was not always correct, their highly confident predictions were mostly correct. These requests, paired with the predicted politeness label, were used to filter crowdworkers. 

\paragraph{Translation Qualification Task}

Inspired by prior work~\cite{pavlick-etal-2014-language} which employed a translation task to assess the language proficiency of crowdworkers, we estimate their language proficiency by evaluating their translation skills.\footnote{In original politeness dataset collection, instead of translation task they introduced paraphrasing task to ensure linguistic attentiveness.}  

We present crowdworkers with a set of five requests (assigned either very polite or very impolite rating by the model) in the target language, and ask them to translate into English as well as to label a politeness score. We first compared the annotator's translation with the output from Google Translate API.\footnote{\url{https://cloud.google.com/translate}} If the edit distance between their translation and the output from Google translate, we remove them from the annotator pool as they could be using this service. We also computed the distance between the user's politeness score and the model's predicted labels, and pruned workers who provided scores that varies significantly from model predictions.

The qualification is not completely automatic, with constant monitoring on four languages on which language-proficient researchers continuously provide sanity checks. Fifteen workers per language took our qualifier task, and after this filtering we ended up with 7 Afrikaans, 9 Spanish, 9 Hungarian, 10 Tamil, 10 Russian, 11 Hindi, 11 Korean, 11 French and 11 Vietnamese workers. 


\paragraph{Final Data Collection / Postprocessing}
The annotators annotated 5 English requests and 15 target language requests instances per task. The annotation interface can be found in the appendix. We collect 3-way annotations for each request. Annotating 20 examples took approximately seven minutes and annotators were paid \$3 for it, translating to \$25.43/hr. 

\subsection{Inter-annotator Agreement}
\label{sec:dataset_metrics}
Ensuring data quality is challenging, especially when we do not have in-house native speaker to inspect for all languages we study. Following prior work~\cite{pavlick-etal-2014-language,danescu-niculescu-mizil-etal-2013-computational}, we estimate the annotation quality by comparing inter-annotator agreement with agreement between randomly assigned labels according to the data distribution\footnote{Specifically, we shuffle the politeness scores of each annotation set (20 examples).}. As we study continuous rather than categorical value, we compute pairwise spearman correlation to measure agreement score instead of Cohen's Kappa.

\begin{table}[t]
    \small
    \centering
    \begin{tabular}{@{}lllll@{}}
    \toprule
    Language     & \multicolumn{2}{c}{en} & \multicolumn{2}{c}{target} \\ \midrule
    hi   & 0.31 & \textit{(0.4)}  & 0.39    & \textit{(0.2)}    \\
    ko   & 0.34 & \textit{(0.34)} & 0.6     & \textit{(0.12)}   \\
    es   & 0.28 & \textit{(0.12)} & 0.52    & \textit{(0.16)}   \\
    ta   & 0.38 & \textit{(0.32)} & 0.33    & \textit{(0.17)}   \\
    fr   & 0.45 & \textit{(0.3)}  & 0.53    & \textit{(0.21)}   \\
    vi   & 0.38 & \textit{(0.31)} & 0.41    & \textit{(0.17)}   \\
    ru   & 0.43 & \textit{(0.34)} & 0.51    & \textit{(0.16)}   \\
    af   & 0.35 & \textit{(0.34)} & 0.37    & \textit{(0.2)}    \\
    hu   & 0.38 & \textit{(0.3)}  & 0.52    & \textit{(0.19)}   \\  \midrule
    average & 0.36 & \textit{(0.31)} & 0.46 & \textit{(0.17)} \\ 
    \bottomrule
    \end{tabular}
    \caption{Pairwise correlation (mean and standard deviation (in brackets)) for each language annotator, on English data and their native language data.}
    \label{tab:irr}
\end{table}

\begin{figure}[t]
    \centering
    \includegraphics[width=0.49\textwidth,trim={0 0 0 0.66cm},clip]{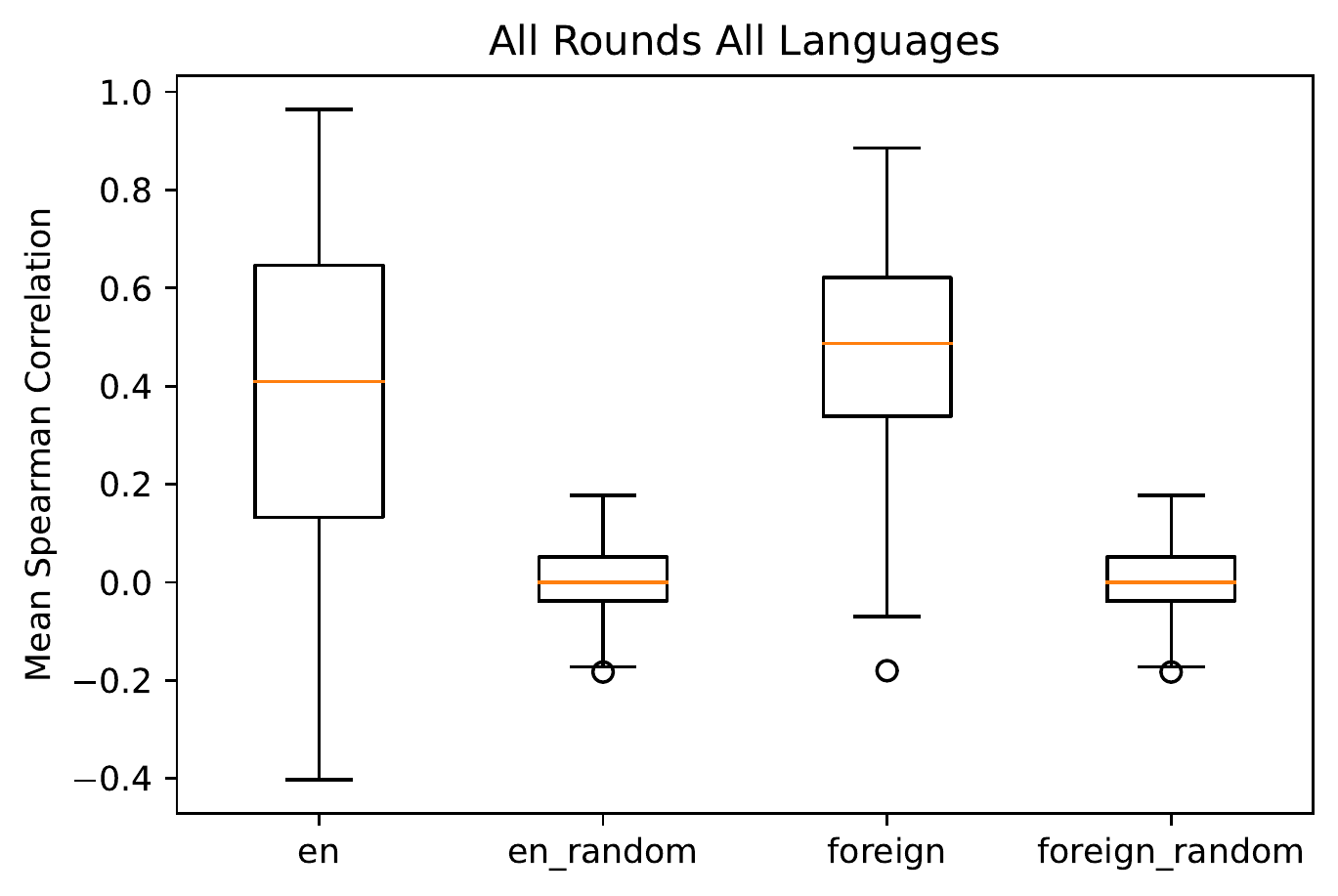}
\vspace{-1.5em}    \caption{Spearman correlation. The fist and third graph represents our annotated data in English and target languages respectively, and the second and the fourth shows correlation for random assignments, which hovers around zero as expected.}\vspace{-0.5em}
    \label{fig:irr_metrics_all}

\end{figure}

As each annotator provided scores for both English sentences and sentences of their native languages, we report both agreement numbers, split by language in Table \ref{tab:irr}. We consistently observe a positive correlation among the annotators' scores. Interestingly, we observe substantially higher agreement when annotators were labeling their own language compared to labeling English across all nine languages. This suggests the interpreting politeness of foreign language can be less precise and more variable compared to interpreting that of native language. As our main goal is collecting target language annotations, this would not impact the quality of our dataset, which studies how native speakers perceive native contents. We plot the averaged pairwise spearman correlation of annotations and that of random assignments in Figure~\ref{fig:irr_metrics_all}. In both English and their native languages, annotator correlation is substantially higher than correlation from random label assignments, which hovers around zero as expected. In Appendix \ref{app:irr_metrics}, we report the correlation with the English politeness labels from the previous study and our annotation, and inter-annotator agreement per by language.
\subsection{Final Dataset}
We collect three way annotations for 500 randomly sampled requests for each language.
\begin{figure}
    \centering
    \includegraphics[width=0.45\textwidth,trim={0 0 0 0.8cm},clip]{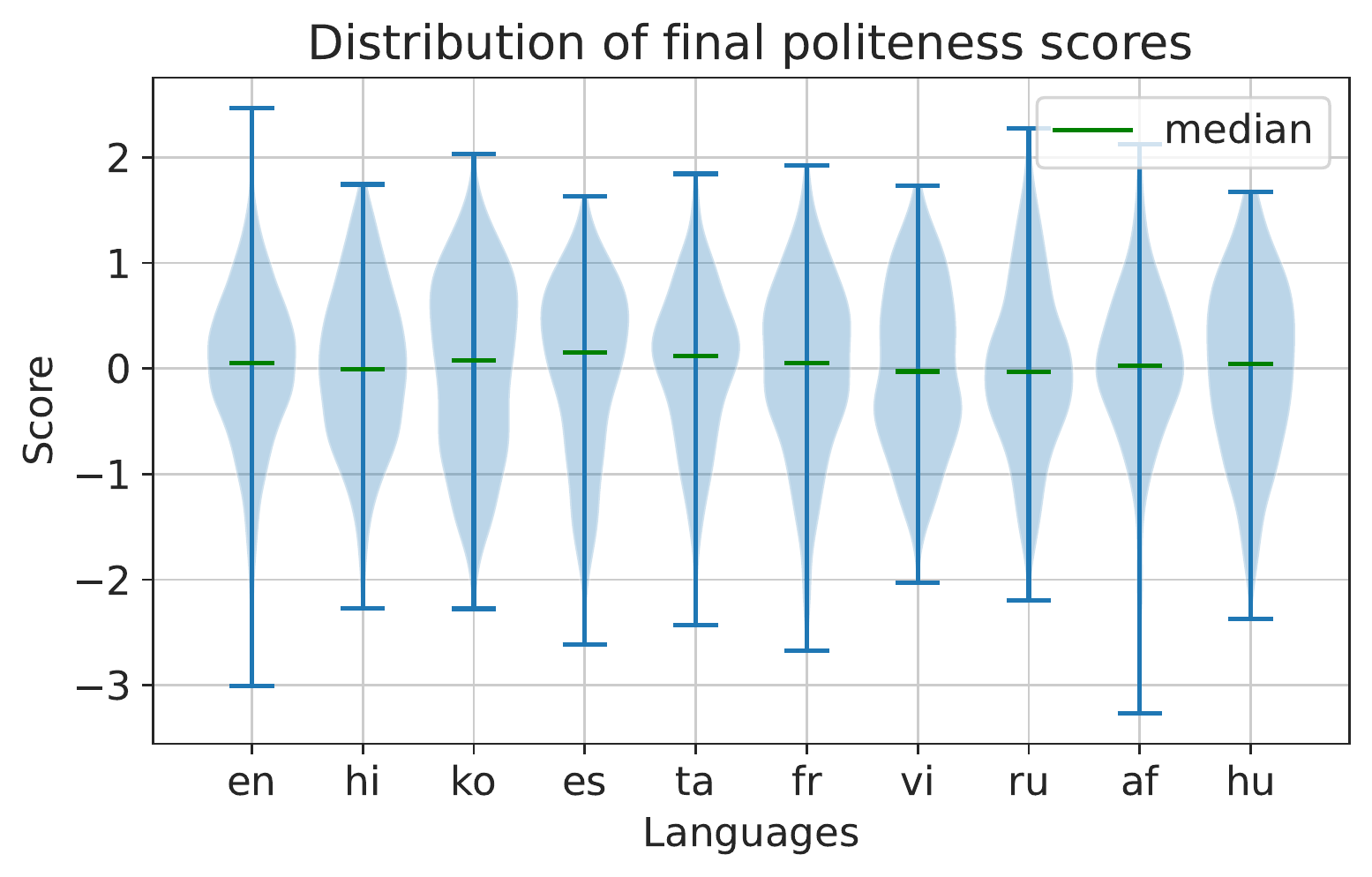}\vspace{-0.5em}
    \caption{Distribution of final politeness scores per language, with mean and median highlighted.}\vspace{-0.5em}
    \label{fig:score_dist}
\end{figure}
We normalize each annotator's score to a normal distribution with a mean of zero and standard deviation of 1, and then average the score of three annotators to get a final score for each item, which ranges from -3 (very impolite) to +3 (very polite). We plot the final politeness distribution per language in Figure~\ref{fig:score_dist}. Examples of annotated sentences are in Appendix \ref{app:request_examples}.

We split these examples into 4 quartiles based on their politeness scores, and consider sentences from the top and bottom 25 percentile of politeness scores only (corresponding to positive and negative politeness), following prior work~\cite{danescu-niculescu-mizil-etal-2013-computational, aubakirova-bansal-2016-interpreting}. This results in a balanced binary politeness prediction task, while reducing the number of examples by half. We refer to this dataset (containing half of the total \data dataset) as \data evaluation dataset. 



\begin{table*}[t]
    \small
    \centering
    \begin{tabular}{@{}llrrrrrrrrrrr@{}}
        \toprule
        Model    & Input Lang.       & en    & hi    & ko    & es    & ta    & fr    & vi    & ru    & af    & hu  &Avg  \\ \midrule
        Majority & -         & 0.537 & 0.5   & 0.5   & 0.5   & 0.5   & 0.5   & 0.5   & 0.5   & 0.5   & 0.5 & 0.5   \\ \midrule
        XLMR & target                 & 0.892 & 0.868 & 0.784 & 0.84  & 0.78  & 0.82  & 0.844 & 0.668 & 0.856 & 0.812 & 0.808 \\
        XLMR   &en  & 0.892 & 0.884 & 0.752 & 0.848 & 0.748 & 0.84  & 0.816 & 0.688 & 0.836 & 0.8 & 0.801   \\
        RoBERTa  &en & 0.912 & 0.868 & 0.692 & 0.836 & 0.768 & 0.812 & 0.796 & 0.684 & 0.856 & 0.768 & 0.786 \\
        \midrule
        GPT3 & target        & 0.808 & 0.732 & 0.708 & 0.732 & 0.596 & 0.764 & 0.692 & 0.688 & 0.688 & 0.76 & 0.706 \\
        GPT3 & en        & 0.808 & 0.668 & 0.62  & 0.732 & 0.652 & 0.72  & 0.664 & 0.612 & 0.7 & 0.652 & 0.668    \\
        \bottomrule
    \end{tabular}
    \caption{Accuracy on \data evaluation dataset. The XLMR and RoBERTa models are finetuned in English politeness data from ~\citet{danescu-niculescu-mizil-etal-2013-computational}, while GPT3 model is prompted in a zero-shot fashion. When Input Lang. column is ``en", we use Google Translate API to translate the target language into English.}
    \label{tab:model_perf}
\end{table*}
\section{Predicting Politeness}\label{sec:classifier}

Equipped with politeness data for nine languages, we evaluate cross-lingual transfer performance of multilingual language models~\cite{conneau-etal-2020-unsupervised}. We are interested in following research questions:
\begin{enumerate}
    \item Can a \textbf{multilingual} model trained on English politeness data predict politeness of different languages?
    \item Can we use a \textbf{monolingual} model trained on English politeness data by translating target languages into English?
\end{enumerate}


\paragraph{Models}
We study two fine-tuned pretrained language models, one English model (RoBERTa~\cite{https://doi.org/10.48550/arxiv.1907.11692}) and one multilingual model (XLM-RoBERTa \cite{conneau-etal-2020-unsupervised}) which supports all nine languages we study. 

We randomly split data from \citet{danescu-niculescu-mizil-etal-2013-computational} to yield 1,926 training and 251 evaluation examples in English. With this training dataset, we fine-tuned each model for five epochs with a batch size of 32 and an learning rate of 5e-6 on a Quadro RTX 6000 machine. We use the large variants for both models. 

At inference time, we translate the target language requests into English using Google Translate API (optional for XLMR model, necessary for RoBERTa model).

We use one large-scale language model, GPT3~\cite{NEURIPS2020_1457c0d6} Davinci-002, in a zero-shot prompting setup with the following prompt:
\begin{verbatim}
    Is this request polite?
    <input example>
\end{verbatim}
Then, we compute the probabilities for two options for next token -- ``yes" and ``no" respectively, which map to ``polite" and ``impolite" labels respectively. Designing prompts for each language is non-trivial, so in this initial study we use this exact same English template for all languages.



\paragraph{Results}


Table \ref{tab:model_perf} reports the model performances. Following recent question answering benchmark~\cite{Clark2020TyDiQA}, we only aggregate the scores on non-English languages to focus on transfer performances. Both finetuned language models (XLMR and RoBERTa) boast strong performance in English, reaching an accuracy hovering 90\%. Even zero-shot GPT model performs competitively, reporting an accuracy of 80.8\%. 

In terms of XLMR model, the results were fairly split on whether it is better to use automatically translated English input, matching the training data, or using the target language input as is. Using the text in English showed better performances in and four (Hindi, Spanish, French, Russian) and using the target language input was better in five languages (Korean, Tamil, Vietnamese, Afrikaans, and Hungarian). Using the target language yields a slightly better performance, questioning whether automatic translation maintain the politeness level. 

Large-scale language model, GPT3, even used in a zero-shot fashion without much prompt engineering~\cite{gao-etal-2021-making} shows competitive performances, significantly outperforming the majority baseline. Similar to XLMR, using target language as is showed better performance than using translated text (70.6 vs. 66.8) on average, and in seven out of nine languages. 



Comparing performances across languages is tricky as the annotation was done by different sets of annotators on different items for each language. To put these numbers in context, we provide a comparison between estimated human performance and model performance in the next section. Would human agreement be lower on languages with weaker model performance?

\begin{figure}
    \centering
    \includegraphics[scale=0.41,trim={0.7cm 0 0 0.66cm},clip]{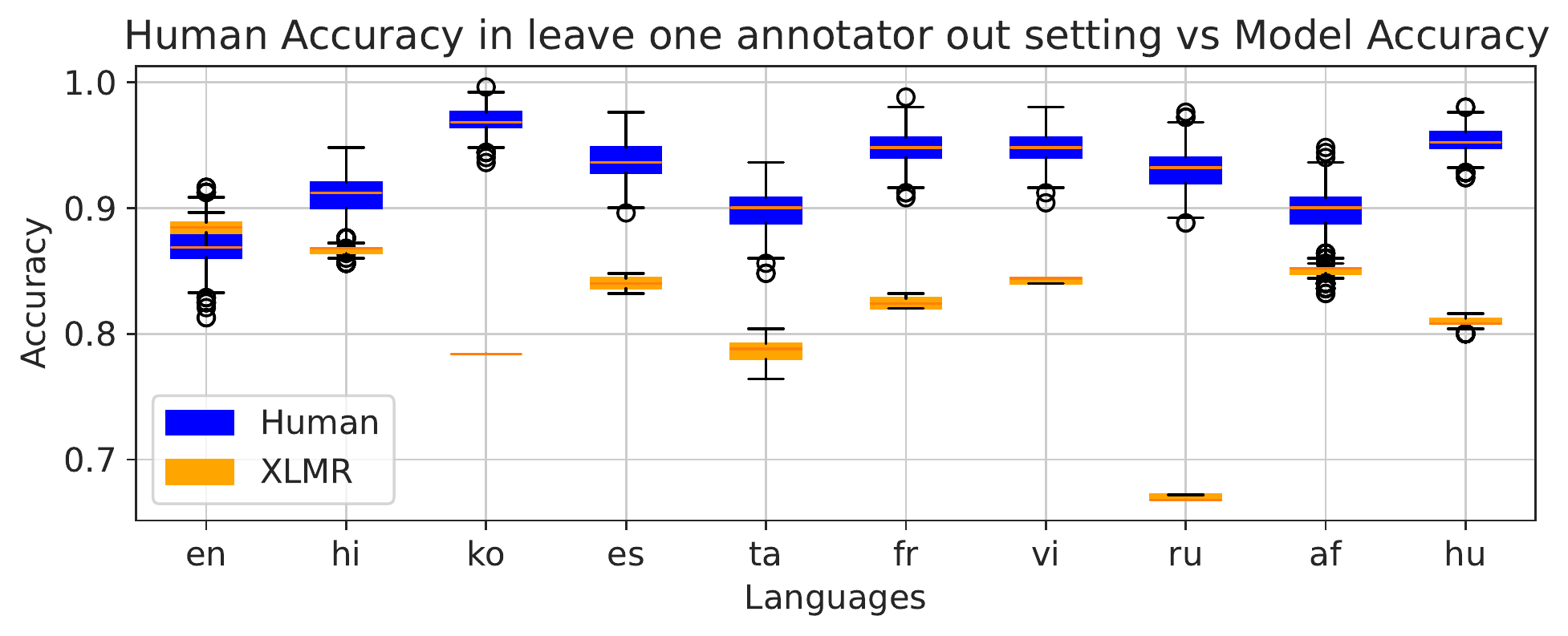}\vspace{-0.5em}
    \caption{Comparing the our best model accuracy (XLMR-target) vs. annotator accuracy on politeness prediction.}\vspace{-1em}
    \label{fig:human_accuracy}
\end{figure}

\paragraph{Comparison with human agreement}
To compute a comparable number between the annotators and models, we use our original 3-way annotated data before aggregating politeness score. We treat one annotator's label as the human prediction and consider the other two as references, taking their mean to get the gold politeness score. We repeat this random sampling process for each example in test set 1,000 times and plot the distribution of accuracy scores in Figure~\ref{fig:human_accuracy}. 


Annotators shows varying degree of agreements -- we notice a particularly stronger agreement in Korean and Hungarian, but overall we observe strong agreement, hovering around 90\%. Interestingly, models significantly underperform in these languages with high human agreement, making the gap between human and model performance large. Six out of the nine languages have a gap of at least 10\%, and two of them being greater than 15\%. 

\section{Building and Analyzing Politeness Strategies in Nine Languages}\label{sec:lexicon}
In this section, we develop a set of linguistic politeness strategies based on existing English strategies \cite{danescu-niculescu-mizil-etal-2013-computational}, and see how can explaining politeness phenomena in nine diverse languages we study.  
While politeness strategies are not necessary for building a high-performing classifier, it can be helpful to understand politeness phenomena. 

The original English study presents a list of politeness strategies along with each strategies relation to assigned politeness score. They found many statistically significant correlations between politeness strategies and human perception, such as words belonging to gratitude lexicon (appreciate), counterfactual modal (could/would) correlates with being polite, and starting the sentence with first person pronoun correlating with being impolite. 

Developing such a politeness lexicon for each language requires expert annotation, which can be infeasible for low-resource languages with a fewer language-proficient researchers~\cite{joshi-etal-2020-state}. Thus, we aim to automatically generate politeness strategies for other languages from the English ones. For this initial study, we focus on lexicon-based strategies (15 out of 20 strategies), excluding strategies involving dependency parsing. 

\paragraph{Mapping English Lexicon to Target Languages}
To build a politeness lexicon in nine languages, we use two NLP tools -- translation and word alignments. 

We sample 5000 Wikipedia editor requests that are not included in our annotated data for each of nine languages.\footnote{For languages (Afrikaans, Hindi) with less than 5K requests, we used all available data. } 
We first automatically translate target language sentence into English (with Google Translate API) and then align the words in the translated English sentence to the words in original sentence in the target language. 

Aligning words in parallel corpora has been long-standing task in NLP. Traditionally, alignments can be obtained as a byproduct of training statistical MT systems \cite{och-ney-2003-systematic,dyer-etal-2013-simple}. Yet, this typically require a large parallel corpus, which we lack for nine languages we study. We instead use alignment method using the similarity between token representations from multilingual pretrained language models (mBERT~\cite{devlin-etal-2019-bert}), fast-align \cite{dou-neubig-2021-word}. 

For each word in English politeness lexicon, we collect their aligned word in the target language. As the alignments maps a sequence of words to a sequence of words, sometimes a single word English lexicon is mapped to multiple words in the target language. For each word in the English lexicon, we consider up to top five target language word sequences as its matching lexicon. We show examples of induced lexicon in Appendix \ref{app:pol_strat} and full lexicon in the repository.

\begin{figure}[t]
    \centering
    \includegraphics[width=0.45\textwidth]{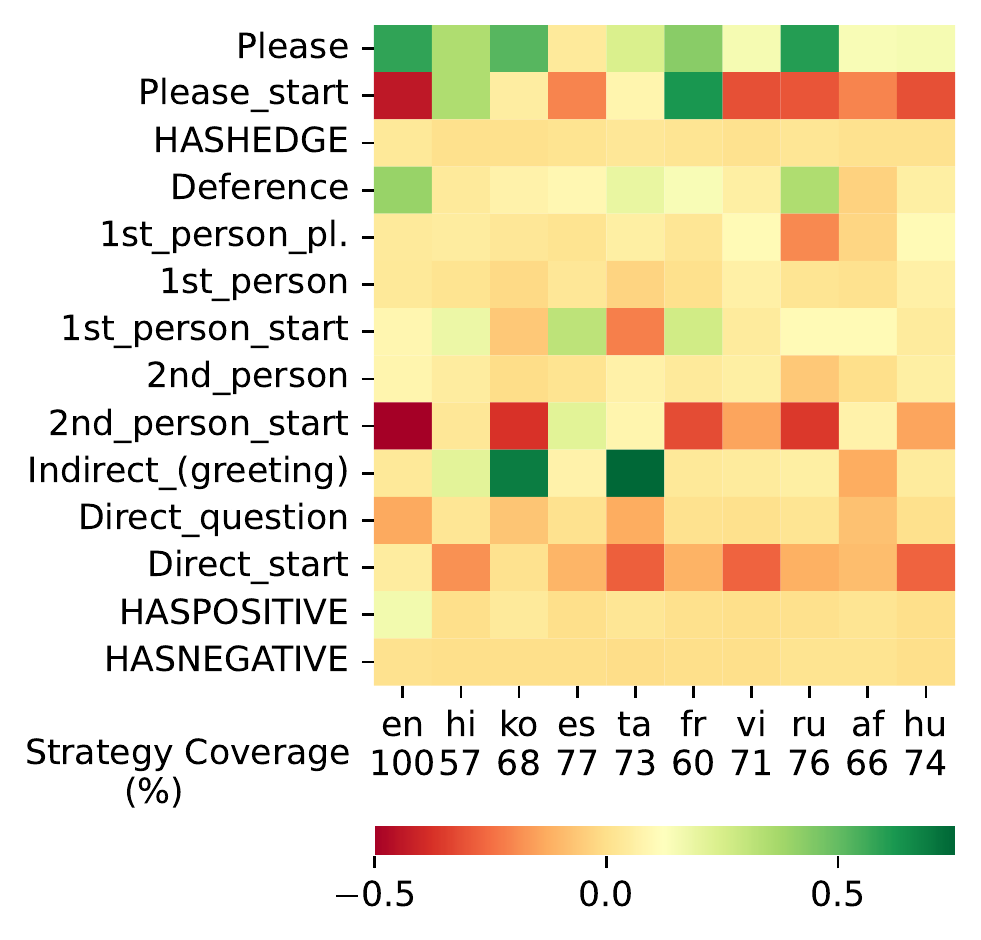}\vspace{-1em}
    \caption{Induced politeness strategies and their relation to politeness scores in nine languages. We plot the average politeness score of a set of sentences containing corresponding strategy. Here, the baseline value is 0. The number of strategies covered by the induced lexicon is also mentioned for each language.}
    \label{fig:avg_score}
\end{figure}

As automatically generated lexicon can be imprecise for either incorrect translation or alignments, we manually inspected the generated lexicon in four languages for which we have language-proficient researchers. We found that the alignments were mostly reasonable, but erroneous and imprecise for words with multiple senses. Not every lexicon was mapped to foreign words either, we show the coverage statistics (average \% of words in lexicon mapped to foreign language words), which hovers around 60-70\%, at the bottom of Figure \ref{fig:avg_score}.

 
\paragraph{Analysis with Induced Lexicon}
Using automatically induced lexicon, we analyze our multilingual politeness data, mirroring the analysis from ~\citet{danescu-niculescu-mizil-etal-2013-computational}. We report the average politeness score of sentences exhibiting each each strategy in Figure \ref{fig:avg_score}. The baseline value here would be 0. We observe that the average politeness score for each strategy across languages are somewhat consistent (e.g., \textsc{Please} strategy being positively correlated in all languages except Spanish). The diverging patterns can be an error with strategy mapping and needs further investigation. Interestingly, in languages with lower model performance (Korean, Tamil), we observe more diverging patterns (e.g., indirect greeting having positive implications in these two languages while mildly negative in English). In the Appendix \ref{app:pol_strat}, we include the occurrence of different strategies in different politeness quartiles (polite or impolite subsets), which exhibits similar pattern. 

\section{Transfer between Formality and Politeness}
\begin{figure}
   
        \small
        \centering
        \begin{tabular}{lrrrr}
        \toprule
        Model                                                              & en    & fr    & it    & pt    \\ \midrule
        Majority Baseline                                                  & \multicolumn{4}{c}{80}     \\
        \midrule
        Before Calib.                                                 & 40.12 & 37.74 & 38.10 & 36.72 \\
        After Calib.                                                  & 73.48 & 74.08 & 74.44 & 73.76 \\
        \bottomrule
        \end{tabular}\vspace{-0.5em}
        \captionof{table}{Transfer from politeness to formality. Formality classification accuracy on X-FORMAL dataset.} 
        \label{tab:formality_results}
    \end{figure}
    \begin{figure}
    
        \small
        \centering
        \begin{tabular}{@{}lllllllllll@{}}
        \toprule
                           & en    & hi    & ko    & es    & ta    \\
                           \midrule
        
        Before Calib. & 0.537 & 0.5   & 0.5   & 0.5   & 0.5   \\
        After Calib.  & 0.557 & 0.612 & 0.588 & 0.644 & 0.564 \\
                            \midrule
                           & fr  & vi    & ru    & af    & hu    \\ \midrule
        Before Calib. & 0.5 & 0.5   & 0.5   & 0.5   & 0.5   \\
        After Calib.  & 0.6 & 0.624 & 0.644 & 0.564 & 0.528 \\ \bottomrule
        \end{tabular}
        \captionof{table}{Transfer from formality to politeness. Politeness classification accuracy on \data evaluation dataset.}
        \label{tab:formality_to_politeness}
\end{figure}

\begin{table}[t]
    \small
    \centering
    \begin{tabular}{p{4cm}cc} 
    \toprule
    Sentence                                                                                                                                 & Formality  & Politeness       \\ 
    \midrule
    Hey, I'm in NYC I'll help you out if your around!                                                                                        & Informal        & {Polite}    \\ 
    I am in New York City. I will help you if you are nearby.                                                                                & Formal          &      Polite                      \\
    \midrule
    why do they try to sound british ?                                                                                                       & Informal        & {Impolite}  \\ 
    Why do they attempt to sound British?                                                                                                    & Formal          &      Impolite                      \\
    \bottomrule
    \end{tabular}
    \caption{Four example with annotated formality label and predicted politeness label. The formality labels are from ~\citet{rao-tetreault-2018-dear} and the politeness labels assigned by our classifier.}
    \label{tab:formality_examples}
    
\end{table}

\begin{table}[t]
    \small
    \centering
    \begin{tabular}{l|r|r}
    \toprule
   \multirow{2}{*}{ Language }  &  $\mathbbm{1}$(polite | formal) =   & p (polite | formal) $>$  \\ 
         &  $\mathbbm{1}$(polite | informal)  &  p (polite | informal) \\ 
    \midrule
    English & 0.811& 0.682  \\
    French  & 0.775& 0.702\\
Italian  & 0.764&0.734\\
Portuguese& 0.779  &0.696 \\
    \bottomrule
    \end{tabular}
    \caption{Analysing politeness predictions on (informal, formal) sentence pairs. The left column represents the fraction of pairs for which the {same politeness label is assigned to both sentences}. The right column represents the fraction of pairs for which the classifier's {probability of being polite for the formal sentence is higher than that of its informal counterparts. }}
    \label{tab:formal_polite}
\end{table}

While we are not aware of computational linguistic studies in politeness covering multiple languages, prior work~\cite{briakou-etal-2021-ola, rao-tetreault-2018-dear} has explored formality in four languages (English, French, Italian and Portuguese). In this section, we study the connections between the formality and politeness. Would formally written sentences perceived as more polite by our classifier? 

We use GYAFC~\cite{rao-tetreault-2018-dear} and X-FORMAL \cite{briakou-etal-2021-ola}, two datasets containing informal sentences from the L6 Yahoo Answers Corpus\footnote{\url{https://webscope.sandbox.yahoo.com/catalog.php?datatype=l}} and four formal rewrites for each sentence (dataset statistics can be found in the Appendix \ref{app:formality}). 

In Table~\ref{tab:formality_results}, we report zero-shot transfer results from politeness classifier to formality classification. We will use our best multilingual politeness classifier (XLMR-target) from Section \ref{sec:classifier}. We calibrate the threshold of our politeness classifier to account for the different data distribution of positive and negative examples. Somewhat surprisingly, the classifier performs worse than the majority baseline. Table \ref{tab:formality_to_politeness} shows performance numbers of transfer in the reverse direction, i.e from formality to politeness. We similarly finetune XLMR model on the English train set from GYAFC~\cite{rao-tetreault-2018-dear}, and evaluated it on \data evaluation dataset, using target language as an input. After the threshold calibration, the model performs better than the majority baseline, but substantially underperforms the in-domain performance reported in Table~\ref{tab:model_perf}.

Does this mean formality and politeness are not linked? Upon inspection (see Table \ref{tab:formality_examples} for examples), we find that politeness prediction for the the informal and formal rewrites of the same sentence often stay consistent. Looking into the model's prediction on (informal, formal) sentence pairs, we find that almost 80\% of pairs in English have the same politeness prediction for both sentences. The left column in Table \ref{tab:formal_polite} depicts this across four languages, suggesting that politeness could be further linked to the \textit{content}, not just \textit{style} of the writing. 

In their original work, \citet{rao-tetreault-2018-dear} report that commonly used techniques to make sentences formal include phrasal paraphrases, punctuation changes, expansions, contractions, capitalization and normalization which are fairly stylistic. Would such rewriting make sentences to be perceived more polite? We investigate this by further looking into (informal, formal) sentence pairs -- for each version of the sentence in the pair, we compute their politeness probability (as assigned by the classifier) and report percentage of pairs where formal version of the same sentence were viewed as more \textit{polite} than its impolite counterpart. The right column in Table~\ref{tab:formal_polite} presents these results -- for about 70\% examples, such rewriting indeed made the sentence perceived as more polite, despite often not enough to flip the politeness decision. 





\section{Related Work}
\paragraph{Politeness \& Formality}
\citet{danescu-niculescu-mizil-etal-2013-computational} presents the first quantitative, linguistic study of politeness, annotating two types of corpora -- requests extracted from conversations between users on Wikipedia User Talk Pages and user comments from Stackoverflow. Followup work explored interpreting neural networks' politeness predictions~\cite{aubakirova-bansal-2016-interpreting} and controllable text generation with target politeness level \cite{sennrich-etal-2016-controlling, niu-bansal-2018-polite,fu-etal-2020-facilitating}. While these work considers politeness phenomena in English, we expand it to study the phenomena in nine languages. A related concept to politeness is formality, studied in multiple prior work~\cite{lahiri2016squinky,pavlick-tetreault-2016-empirical,rao-tetreault-2018-dear,briakou-etal-2021-ola}.




\paragraph{Multilingual Models}
Recent progresses in pretrained language models have brought better representation for multitude of languages. Multilingual language models like mBERT \cite{devlin-etal-2019-bert}, XLMR \cite{conneau-etal-2020-unsupervised}, based on the transformer architecture, are pretrained with the masked language modeling objective on a large amount of corpora~\cite{ElKishky2020AMC,OrtizSuarez2019AsynchronousPF} spanning over 100 languages. While the community also recognizes the varying quality of unlabeled data in a range languages~\cite{Caswell2022QualityAA}, such multilingual models provide improved representations for modeling low resource languages. When finetuned on downstream task data in a single language, these models make reasonable predictions in multiple languages~\cite{Wu2019BetoBB}. Multilingual models have also been evaluated in a prompting setup for different tasks like Machine Translation \cite{tan-etal-2022-msp} and different Multilingual NLU tasks \cite{zhao-schutze-2021-discrete, https://doi.org/10.48550/arxiv.2112.10668, winata-etal-2021-language}.

\paragraph{Multilingual Benchmarks}
Despite recent progresses in NLP resources and benchmarks, partially powered by affordable crowdsourcing~\cite{Snow2008CheapAF}, linguistic resources in low resource languages are still severely limited to compared to resources in English~\cite{joshi-etal-2020-state}. Many existing datasets are translated from English data~\cite{conneau-etal-2018-xnli,Longpre2021MKQAAL}. While translating approach for dataset construction have advantage of ensuring similar data distribution across languages, data collected in such fashion will not reflect the language usages of diverse population, introducing translationese which can be different from purely native text~\cite{Lembersky2011LanguageMF}. We provide resources for nine typologically diverse languages, capturing a subtle phenomena of politeness. 

\section{Conclusion}
We present \dataset, a corpus of requests paired with its perceived politeness score spanning nine languages. We evaluate multiple multilingual models in zero-shot politeness prediction and find that they are able to perform well without being trained on data from the same language, while not reaching human-level performances yet. 


\section*{Limitations}
Our dataset is moderately sized (250 examples per language in the evaluation portion, and a total of 500 examples per language) and still covers a limited number of languages. We had intended to cover more languages (one example being Japanese), but this were hindered by the number of annotators we could recruit for each language. 

The aligned politeness strategy lexicon (Section~\ref{sec:lexicon}) relies on multiple automatic toolkits (machine translation system and word alignments), thus analysis should be interpreted with caution.  



\section*{Ethical Considerations}
The data we annotate comes from Wikipedia User Talk pages, which is an online forum for communication between editors on Wikipedia. This data spans nine different languages and contains speakers from different countries and demographics. The annotation is done by crowdworkers recruited from the online platform Prolific. These workers aren't restricted to a particular country. They are paid a wage of \$25.43/hr which is higher than the average pay stipulated on the platform. We use this data to evaluate an existing model across multiple languages, and do not use it for training as such.

\section*{Acknowledgements}
We would like to thank Yasumasa Onoe, Bernardo Oviedo, Gokul Anandaraman for their help in the earlier phase of the project development and inspecting data quality. We would also like to thank Cristian Danescu-Niculescu-Mizil for answering questions about his work. We'd like to thank Joel Tetreault and Yahoo for providing access to the formality datasets. We also thank Akari Asai for providing feedback on the paper. We'd like to thank Anuj Diwan for the thoughtful discussions on this topic and providing helpful feedback along the way.

\bibliography{anthology,custom}
\bibliographystyle{acl_natbib}

\clearpage
\appendix
\section{Annotation UI}
\begin{figure*}
    \centering
    \includegraphics[width=\textwidth]{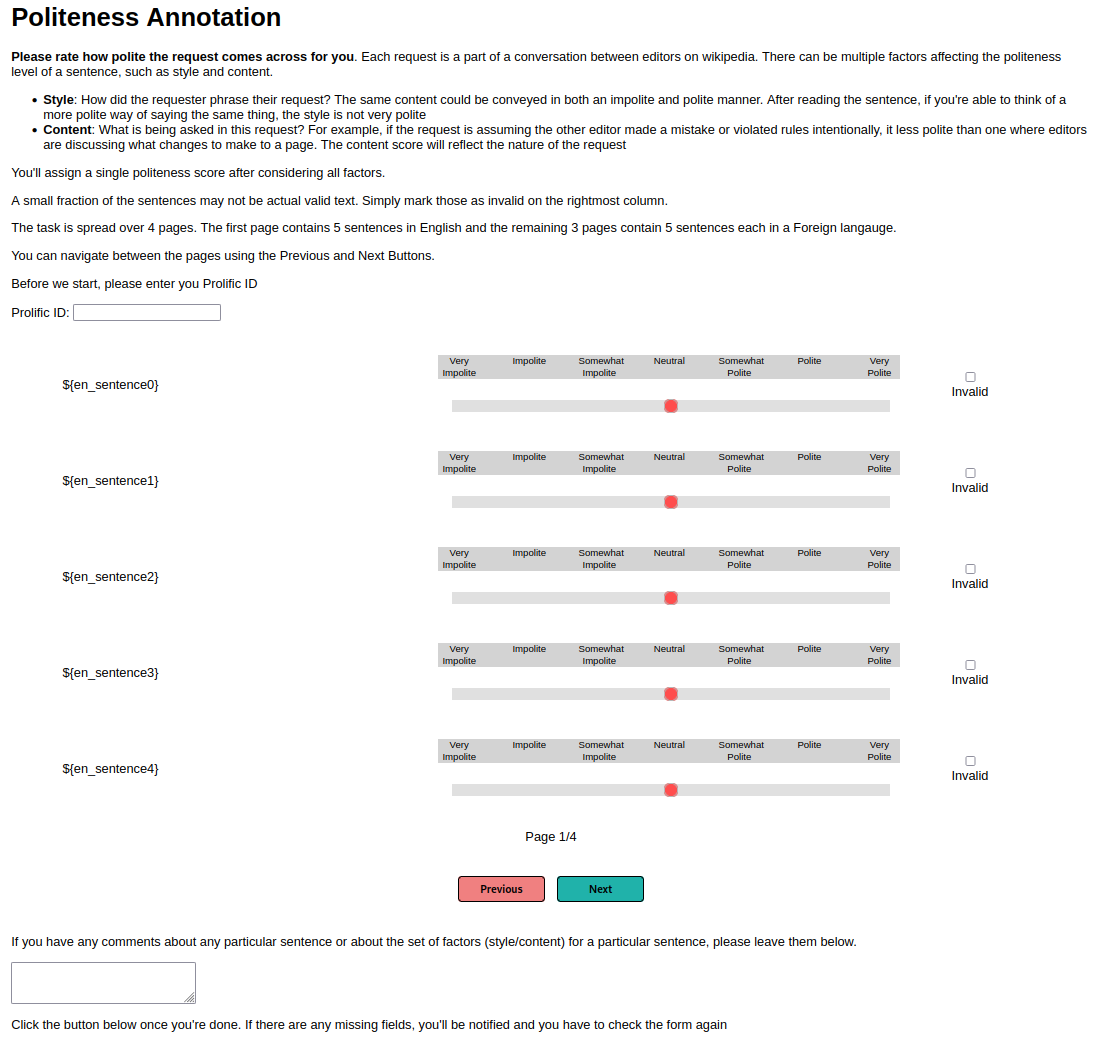}
    \caption{Annotation Interface}
    \label{fig:annotation_ui}
\end{figure*}
Figure \ref{fig:annotation_ui} contains the user interface used for the final annotation process.

\section{Example Requests}
\label{app:request_examples}
\begin{figure*}
    \centering
    \includegraphics[width=0.999\textwidth]{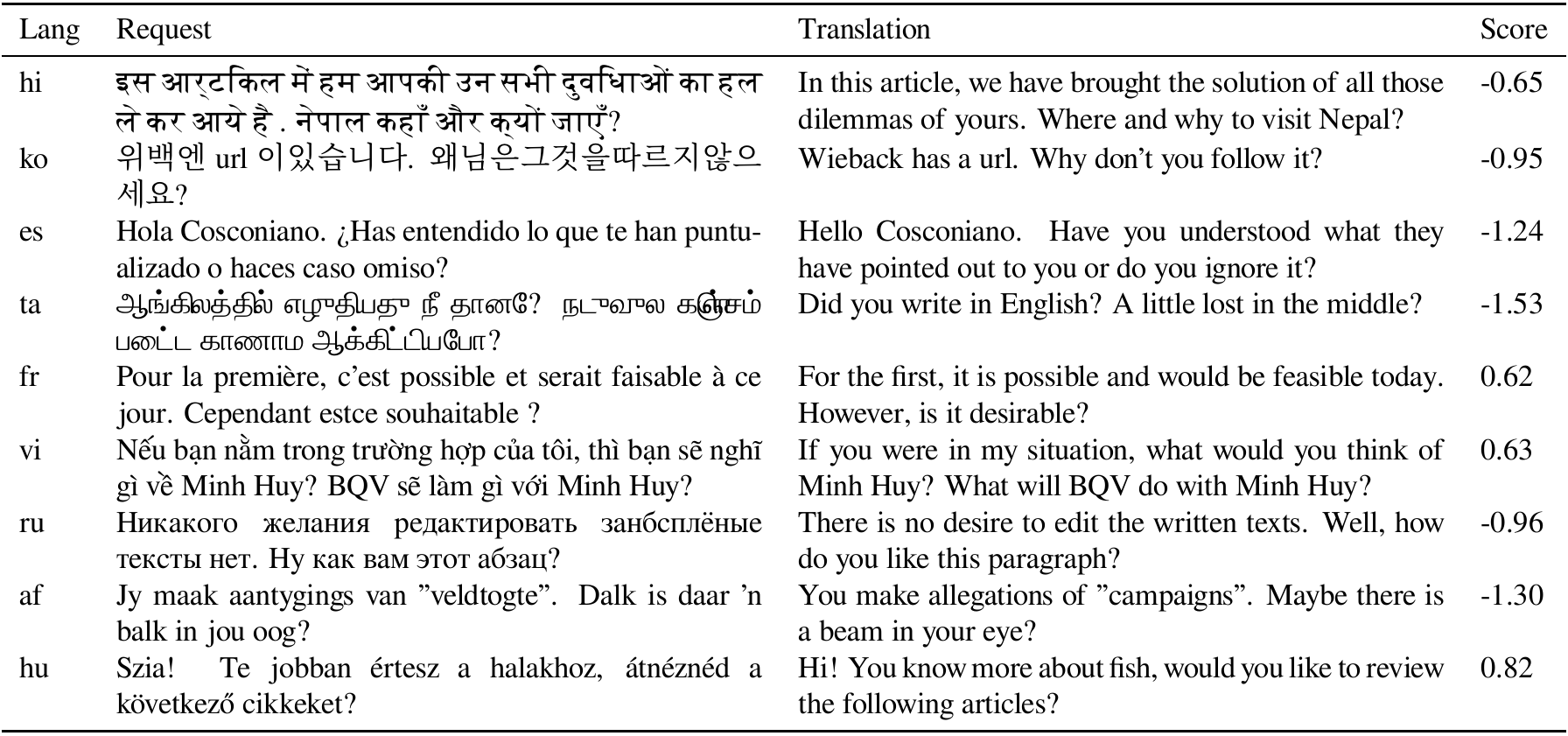}
    \captionof{table}{Examples from \data dataset. The politeness scale is from -3 (very impolite) to +3 (very polite). }
    \label{tab:request_examples}
\end{figure*}

Table \ref{tab:request_examples} contains examples of requests in different languages and the politeness score assigned to them.

\section{Additional Inter Annotator Agreement Reports}
\label{app:irr_metrics}
\begin{figure}[t]
    \centering
    \includegraphics[width=0.49\textwidth]{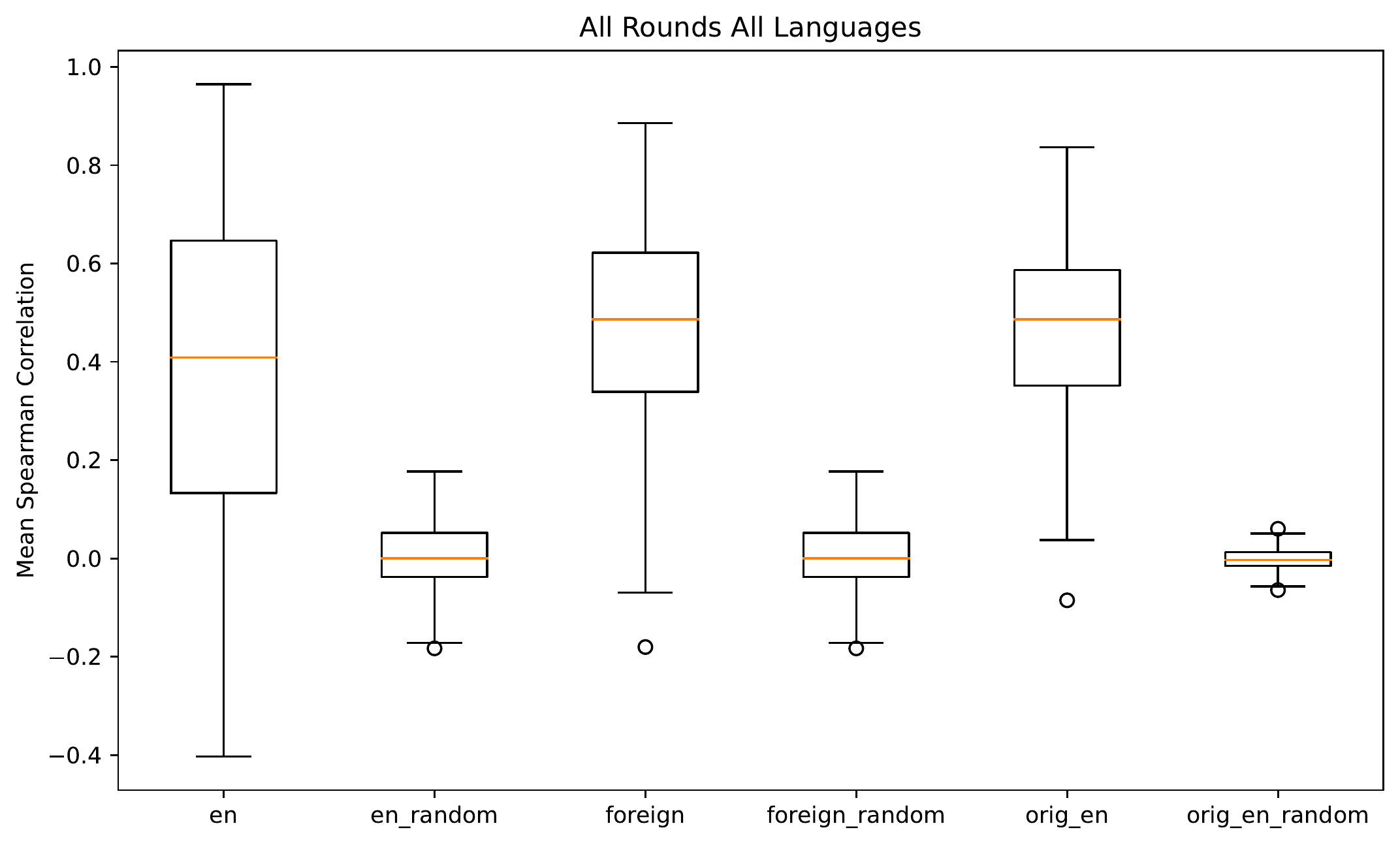}
    \caption{Pairwise correlation metric on our annotations compared to the annotations released by \citet{danescu-niculescu-mizil-etal-2013-computational}}
    \label{fig:irr_new_vs_old}
\end{figure}

Figure \ref{fig:irr_new_vs_old} compares the overall IRR metrics on our annotations with the IRR on the annotations released by \citet{danescu-niculescu-mizil-etal-2013-computational} on English request data. They release the 5 way annotation done on their data and also a single score for each sentence after averaging and normalization. We report two scores in Table \ref{tab:irr_2}: a correlation with raw annotations and a correlation with the final aggregated scores. 

\begin{table}[t]
\centering
    \small
    \centering
    \begin{tabular}{lllll}
    \toprule
         & \multicolumn{2}{c}{Raw Score } & \multicolumn{2}{c}{Aggregate Score} \\ \midrule
    hi   & 0.27      & \textit{(0.48)}      & 0.41      & \textit{(0.44)}     \\
    ko   & 0.27      & \textit{(0.48)}      & 0.45      & \textit{(0.46)}     \\
    es   & 0.24      & \textit{(0.48)}      & 0.4       & \textit{(0.42)}     \\
    ta   & 0.25      & \textit{(0.48)}      & 0.49      & \textit{(0.39)}     \\
    fr   & 0.25      & \textit{(0.46)}      & 0.52      & \textit{(0.35)}     \\
    vi   & 0.29      & \textit{(0.48)}      & 0.46      & \textit{(0.39)}     \\
    ru   & 0.21      & \textit{(0.52)}      & 0.45      & \textit{(0.4)}      \\
    af   & 0.28      & \textit{(0.48)}      & 0.44      & \textit{(0.44)}     \\
    hu   & 0.22      & \textit{(0.5)}       & 0.48      & \textit{(0.38)}    \\ \bottomrule
    \end{tabular}
    \caption{Agreement with original English labels \cite{danescu-niculescu-mizil-etal-2013-computational} annotated by speakers of various languages. Mean and std deviation (in brackets).}
    \label{tab:irr_2}
\end{table}

\begin{figure*}[t]
    \centering
    \includegraphics[width=\textwidth]{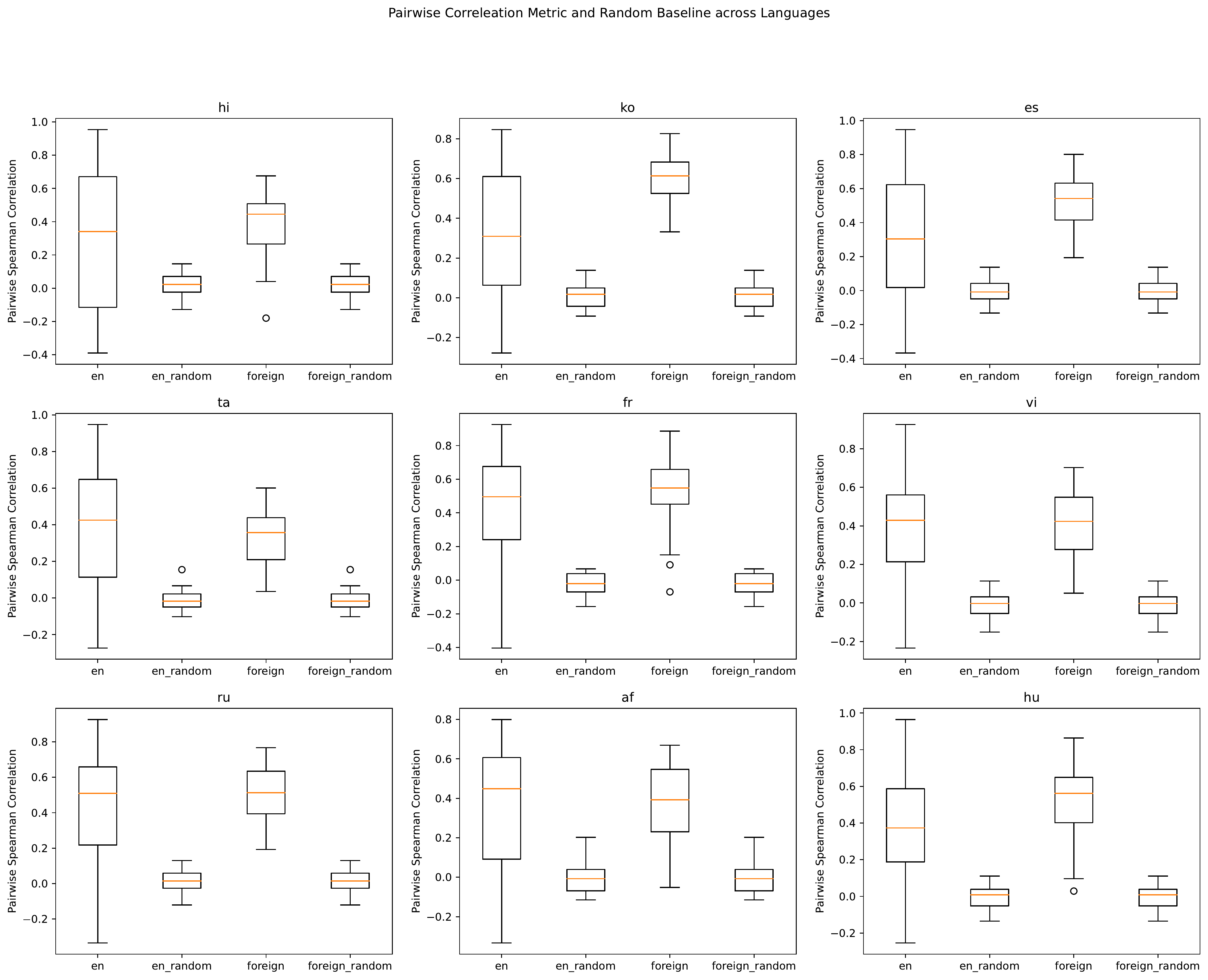}
    \caption{Pairwise Correlation Metric over different HITs, along with Random Baseline value for different Languages}
    \label{fig:irr_boxplots}
\end{figure*}

Figure \ref{fig:irr_boxplots} shows the distribution of the pairwise correlation metric over different HITs for each language. Each subplot has the distribution over the english and target language parts of each HIT, as well as a baseline method where the scores are shuffled before computing the correlation.

The correlations in the random baseline are close to 0 and the correlations on the annotations are significantly higher. The correlations on the English annotations do show more variance in their distribution.

\section{Politeness Score Statistics}
\begin{table*}[t]
    \centering
    \begin{tabular}{@{}lllllll@{}}
    \toprule
    Lang      & \# of       & \multicolumn{4}{c}{Scores}                                 & \% \\
    \textbf{} & Examples & \textit{Mean} & \textit{Std} & \textit{Min} & \textit{Max} & Positive \\ \midrule
    hi        & 500       & -0.0005       & 0.7756       & -2.2745      & 1.7437       & 0.4980   \\
    ko        & 500       & -0.0016       & 0.8659       & -2.2760      & 2.0335       & 0.5320   \\
    es        & 500       & -0.0007       & 0.8241       & -2.6132      & 1.6320       & 0.5740   \\
    ta        & 500       & 0.0035        & 0.7546       & -2.4319      & 1.8454       & 0.5540   \\
    fr        & 500       & 0.0003        & 0.8406       & -2.6700      & 1.9268       & 0.5340   \\
    vi        & 500       & -0.0019       & 0.7810       & -2.0258      & 1.7338       & 0.4920   \\
    ru        & 500       & 0.0047        & 0.8234       & -2.1938      & 2.2763       & 0.4880   \\
    af        & 500       & 0.0049        & 0.7845       & -3.2678      & 2.1264       & 0.5200   \\
    hu        & 500       & 0.0027        & 0.8322       & -2.3750      & 1.6705       & 0.5200   \\ \bottomrule
    \end{tabular}
    \caption{Statistics on Final Politeness Scores}
    \label{tab:data_summary}
\end{table*} 
Table \ref{tab:data_summary} summarizes the distribution of scores across languages. All the languages have a mean close to 0, with similarly shaped distribution of scores. The minimum and maximum scores seem to vary a bit across languages. Some languages like Spanish have a higher median score and a higher number of sentences with a positive scores.

\section{Politeness Strategies}
\label{app:pol_strat}
\begin{figure*}[t]
    \centering
    \includegraphics{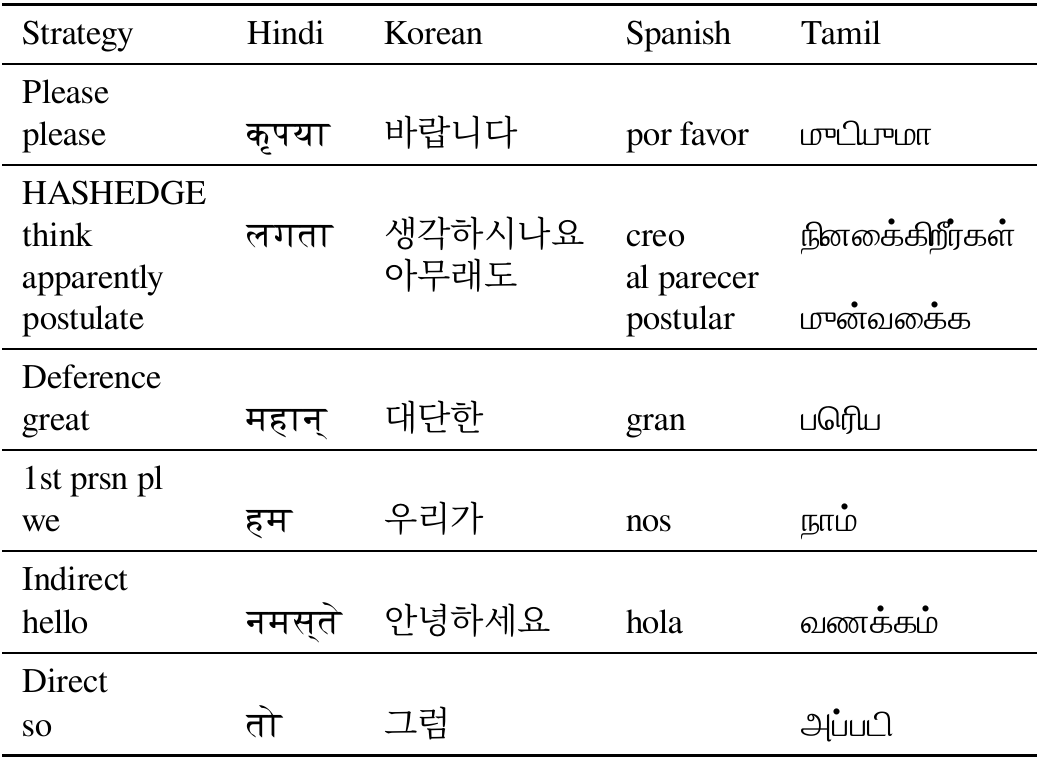}
    \captionof{table}{Examples of politeness strategies lexicon gathered from our alignment method and then cherry picked by language proficient researcher.}
    \label{tab:pol_strat_examples}
\end{figure*}
Table \ref{tab:pol_strat_examples} gives some examples of the politeness strategy lexicon we obtained by our automated method.

\section{Politeness Strategy Distribution}
\begin{figure*}[t]
    \begin{subfigure}{0.45\textwidth}
        \includegraphics[width=1\textwidth]{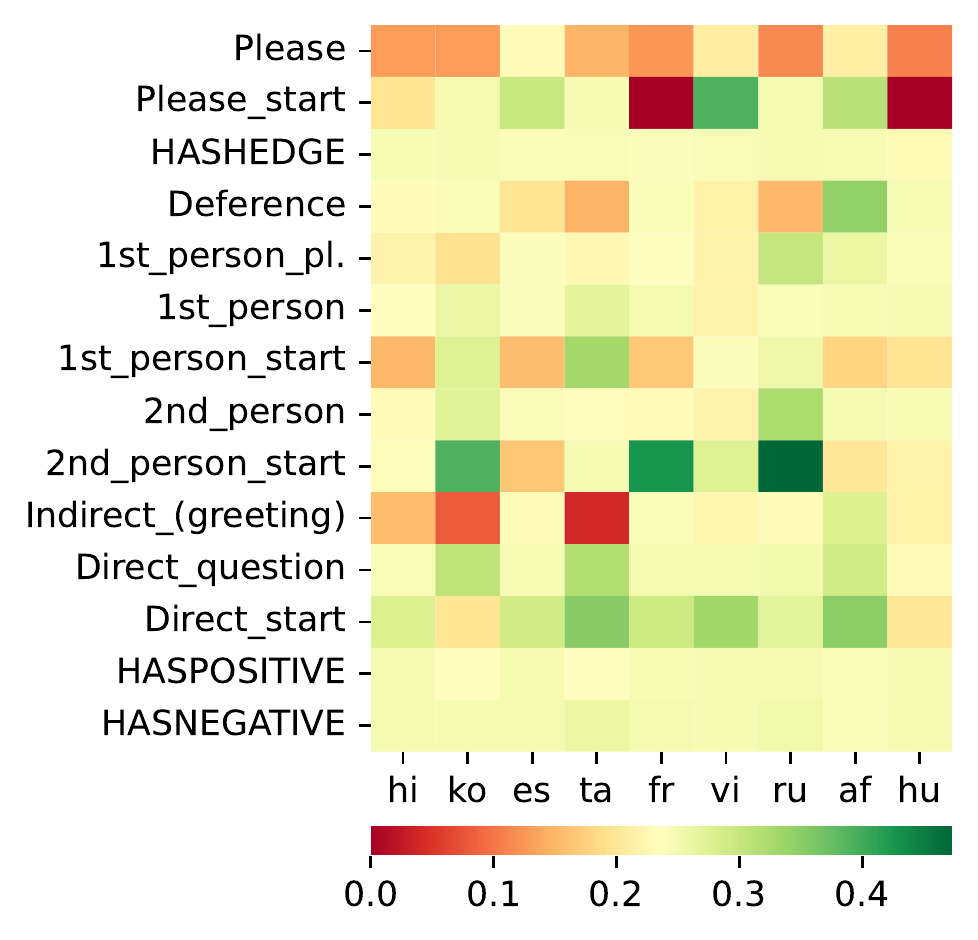}    
    \end{subfigure}
    \hfill
    \begin{subfigure}{0.45\textwidth}
        \includegraphics[width=1\textwidth]{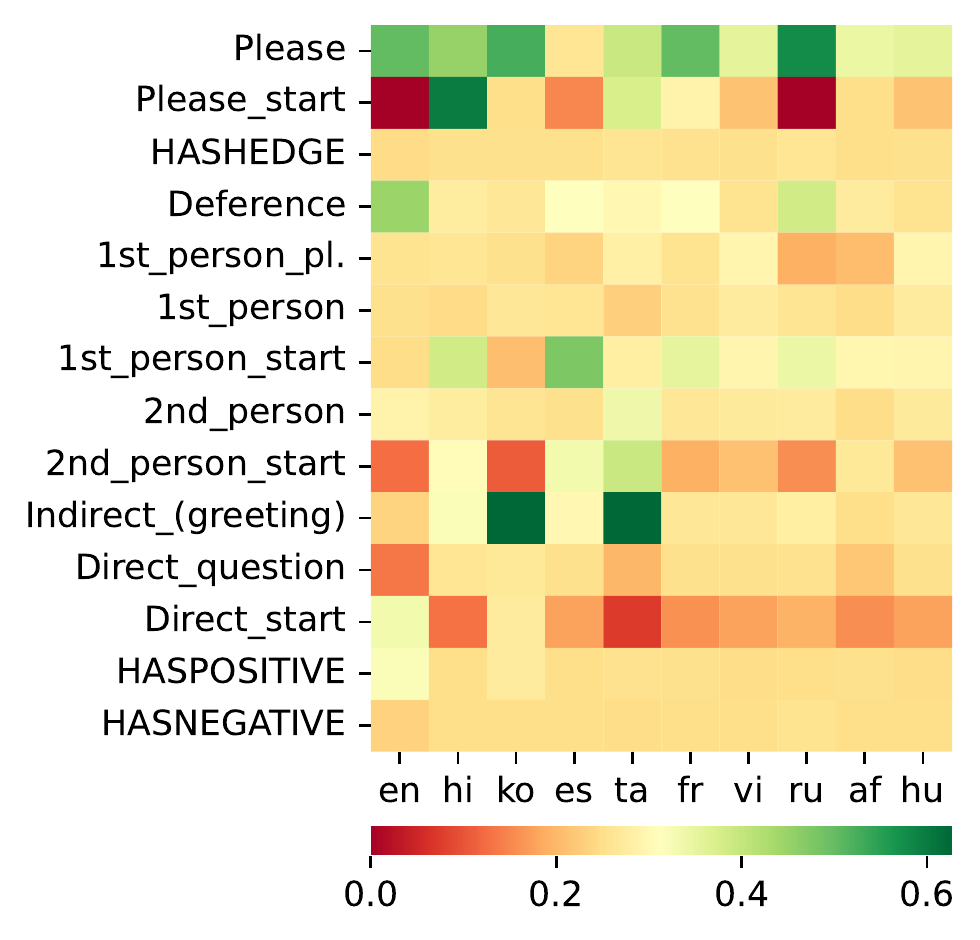}
    \end{subfigure}\vspace{-1em}
  \caption{Presence of Politeness Strategies across the 1st and 4th quartiles (left and right plot) of data. Any cells deviating from the baseline value of 0.25 represent significant results} 
  \label{fig:strategy_distribution}
\end{figure*}

Figure \ref{fig:strategy_distribution} showcases the occurrence of strategies in sentence belonging to the least polite (1st quartile) and most polite (4th quartile) subsections of our data. Cells shaded in light orange represent a baseline value of 0.25 and anything deviating from this appear in Dark Green or Red. We can clearly see difference across the the 2 quartiles for some of these strategies.

\section{Politeness to Formality Transfer}
\label{app:formality}
\begin{table}[t]
    \small
    \centering
    \begin{tabular}{lll}
    \toprule
    Dataset    & \# Informal & \# Formal \\ \midrule
    \citet{rao-tetreault-2018-dear} \\
    English    & 2478        & 10992     \\
    \midrule
    \citet{briakou-etal-2021-ola} \\
    French     & 1000        & 4000      \\
    Italian    & 1000        & 4000      \\
    Portuguese & 1000        & 4000      \\ \bottomrule
    \end{tabular}
    \caption{Statistics of Formality Data used for Evaluation (as test sets)}
    \label{tab:formality_stat}
\end{table}

We use the XLMR classifier trained in Section \ref{sec:classifier} and evaluate it on the mix of informal and formal sentences (1:4 ratio) as a test set. These performance numbers are shown in Table \ref{tab:formality_results}. We report the classifier Accuracy, as well as a majority baseline. Since we have an imbalanced mix of sentences, we decided to calibrate the classifier's threshold using the dev set. We get the probability for the 80th percentile of scores from the English dev set and use this on the test sets.

\end{document}